\documentclass[sigplan,screen]{acmart}

\AtBeginDocument{%
  \providecommand\BibTeX{{%
    \normalfont B\kern-0.5em{\scshape i\kern-0.25em b}\kern-0.8em\TeX}}}

\setcopyright{acmcopyright}
\copyrightyear{2022}
\acmYear{2022}
\acmDOI{XXXXXXX.XXXXXXX}

\acmConference[SIGKDD 2022]{KDD ’2022: The 28th ACM SIGKDD International
Conference on Knowledge Discovery \& Data Mining}{June August 14-18,
  2022}{Washington, DC}
\acmPrice{15.00}
\acmISBN{978-1-4503-XXXX-X/18/06}
\usepackage{graphicx}
\usepackage{subcaption}
\usepackage{enumitem}
\usepackage{amsmath}
\usepackage{breqn}
\usepackage{lipsum}
\usepackage{multirow}
\usepackage[normalem]{ulem}
\usepackage{caption}
\usepackage{tabu}
\usepackage{multicol}
\usepackage{multibib}

\begin{document}

\title{SAMCNet for Spatial-configuration-based Classification:  A Summary of Results}


\author{\{Majid Farhadloo, Carl Molnar,  Gaoxiang Luo, Yan Li, Shashi Shekhar\}}
\authornote{\{farha043, molna018, luo00042, lixx4266 , shekhar\}@umn.edu.}

\affiliation{%
    \institution{Department of Computer Science and Engineering, University of Minnesota}
        \country{USA}
 }

\author{\{Rachel L. Maus, Svetomir N. Markovic\}}
\authornote{\{maus.rachel, markovic.svetomir\}@mayo.edu.}
\affiliation{%
  \institution{Department of Immunology, Mayo Clinic}
    \country{USA}
}
\author{\{Alexey Leontovich, Raymond Moore\}}
\authornote{\{leontovich.alexey, moore.raymond\}@mayo.edu.}
\affiliation{%
\institution{Department of Quantitative Health Sciences, Mayo Clinic}
    \country{USA}
}

\renewcommand{\shortauthors}{Farhadloo, et al.}


\begin{abstract}
The goal of spatial-configuration-based classification is to build a classifier to distinguish two classes (e.g., responder, non-responder) based on the spatial arrangements (e.g., spatial interactions between different point categories) given multi-category point data from two classes. This problem is important for generating hypotheses in medical pathology towards discovering new immunotherapies for cancer treatment as well as for other applications in biomedical research and microbial ecology. This problem is challenging due to an exponential number of category subsets which may vary in the strength of spatial interactions. Most prior efforts on using human selected spatial association measures may not be sufficient for capturing the relevant (e.g., surrounded by) spatial interactions which may be of biological significance. In addition, the related deep neural networks are limited to category pairs and do not explore larger subsets of point categories. To overcome these limitations, we propose a Spatial-interaction Aware Multi-Category deep neural Network (SAMCNet) architecture and contribute novel local reference frame characterization and point pair prioritization layers for spatial-configuration-based classification. Extensive experimental results on multiple cancer datasets show that the proposed architecture provides higher prediction accuracy over baseline methods.



\end{abstract}



\maketitle

\section{Introduction} 

\begin{figure}
    \centering
    \includegraphics[width=.7\linewidth, height = 5cm]{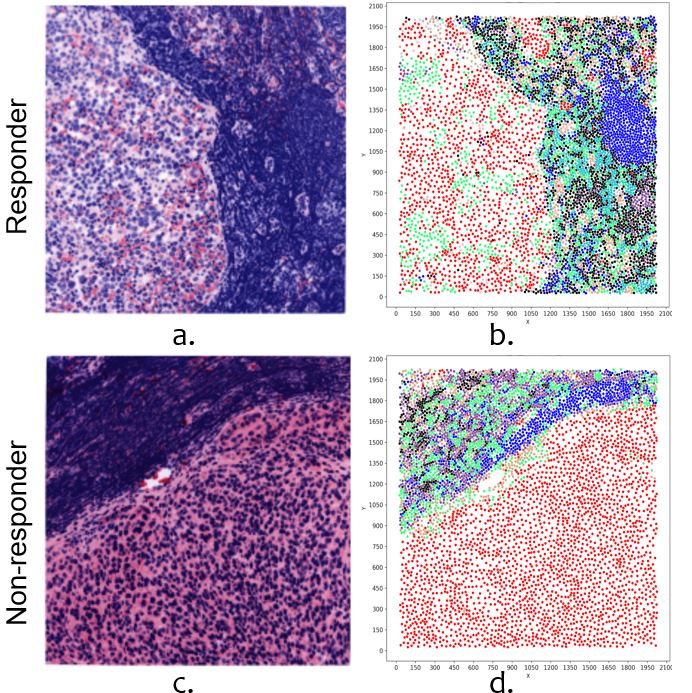}
    \caption{(a,c). Virtual H\&E images of responder and non-responder at the tumor-margin, (b,d)
    corresponding point patterns detailing all cell classes.}
    \label{example1}
    \vspace*{-2\abovedisplayskip}
\end{figure}

Spatial-configuration-based classification aims to build a classifier that can learn spatial patterns in multi-category point patterns to distinguish between two classes. When each data point belongs to a distinct category feature, it logically follows that the value of different spatial interactions between various points is additionally significant for the classification task. For example, the impact of one type of immune cells (e.g., cytotoxic T lymphocytes (CTLs)) on nearby cancer cells may be affected by other immune cells (e.g., T regulatory cells) \cite{wang2009pd1}.  A multi-category point pattern is a collection of spatially-defined objects with corresponding categorical features (e.g., immune and tumor cells). Fig. \ref{example1} illustrates the point patterns of the pathologist-driven field of views (FOV) of human tissue samples stained with a chemical dye (e.g., hematoxylin/eosin (H\&E)) at the tumor-margin between two classes of responder and non-responder indicating two clinical response to immunotherapy treatment. Each data point shows the centroid coordinates of a cell in a pixel and the color of its corresponding cell type. 



This problem is important because spatial configurations, a proxy for physical interactions, help generate new hypotheses towards discovering disease therapeutics (e.g., immunotherapies for cancer treatment). These hypotheses could be used in applications such as medical pathology, biomedical research, and microbial ecology, where multi-category point patterns frequently appear. In many diseases, such as cancer, spatial arrangement and associations between distinct phenotypic markers are crucial to understand normal tissue function and disease biology. For example, the development of effective intervention strategies relies on the knowledge of the spatial arrangement and cellular mechanisms of coronavirus infections \cite{v2021coronavirus}. Researcher in immunology also seeks to understand the spatial configuration of immune and tumor cells to help evaluate the effectiveness of the immune checkpoint inhibitors (e.g., anti-PD-1) in cancer treatment \cite{gide2020close}. These indicate the need for an automated system for analyzing spatial interactions at the molecular level to identify targets for disease therapeutics. Table \ref{use_case} presents use case of the proposed model in different application domains.

This problem is challenging due to the following reasons. First, the number of potential spatial patterns is exponentially related to the number of different category subsets. In addition, the spatial association between various point pair instances is not always equal, requiring the model to learn these distinctions. CTL-tumor cell interactions, for example, are more biologically relevant in the context of effector function (e.g., a CTL must engage a tumor cell to kill it); in contrast, B cell-tumor cell interactions are more indirect \cite{farhood2019cd8}. A point pair instance describes the spatial relationship between a point belonging to one category and its neighbor belonging to a same or different  category. 
The second challenge is that multi-category point patterns are heterogeneous and form complicated structural and higher-order spatial interactions. Lastly, point patterns possess different properties (e.g., invariance to permutation), which means the classifier needs to meet the same requirements to achieve a robust surrogate representation.


Most of the prior works to identify spatial associations in multi-category point patterns can be classified into hand-constructed features using spatial association interest measures (e.g., Pearson correlation, G-cross, Ripley’s cross-k, Participation index) \cite{maley2015ecological, barua2018spatial, schapiro2017histocat, lu2021feature, yan2019understanding}. For example, \cite{maley2015ecological} uses the classical statistics measures (e.g., Pearson Correlation), which are sensitive to the choice of spatial partitioning. In addition, neighbor-graph-based approaches (e.g., \cite{shekhar2001discovering, barua2018spatial}) are primarily a function of distance, which may not accurately model the true spatial relationships among categorical points on the three-dimensional surface (e.g., organ). Furthermore, these techniques are used in isotropic space, with the same intensity regardless of measurement direction, which may not be enough to capture relevant features that might be biologically significant. Lastly, analyzing the spatial association between different communities (e.g., tumor and stromal cells indicated as sub-graph-level) \cite{lu2021feature, yan2019understanding} does not reveal critical information about the spatial relationships between distinct categorical points within each community. In more recent work, a spatial-relationship aware neural network (SRNet) \cite{li2021srnet} aims to confound these limitations leveraging machine-constructed features to model spatial relationships between points of different categories. However, SRNet is limited to only binary spatial relationships, and the importance between distinct binary category pairs is assumed to be equal.

\begin{table}\scriptsize
\centering
\caption{Use cases of proposed model}
\begin{tabular}{|l|l|} 
\hline
Application Domain   & Use Case (Objective) \\ 
\hline
Medical Pathology    & \begin{tabular}[c]{@{}l@{}}Understanding the spatial configuration between~immune\\~ and tumor cells to evaluate the effectiveness of ICI \cite{gide2020close}\end{tabular}  \\ 
\hline
Pharmacology              & \begin{tabular}[c]{@{}l@{}}Identifying protein interactions and bindings\\~towards discovering structure-based drugs \cite{farag2020identification} \end{tabular}  \\ 
\hline
Ecology              & Inferring predator-prey spatial interactions in food webs \cite{pomeranz2019inferring} \\ 
\hline

Paleontology & \begin{tabular}[c]{@{}l@{}}Studying fossils to classify organisms and examine their \\  interactions with each other and the environment \cite{garwood2019re} \end{tabular} \\ 
\hline
Epidemiology         & \begin{tabular}[c]{@{}l@{}}Investigating the relationship between human mobility\\and spread of Covid-19 \cite{sharma2022understanding}\end{tabular}  \\ 
\hline
\end{tabular}
\label{use_case}
\vspace*{-\abovedisplayskip}
\end{table}

\begin{figure}
  \begin{subfigure}[b]{1\columnwidth}
    \includegraphics[width=\linewidth]{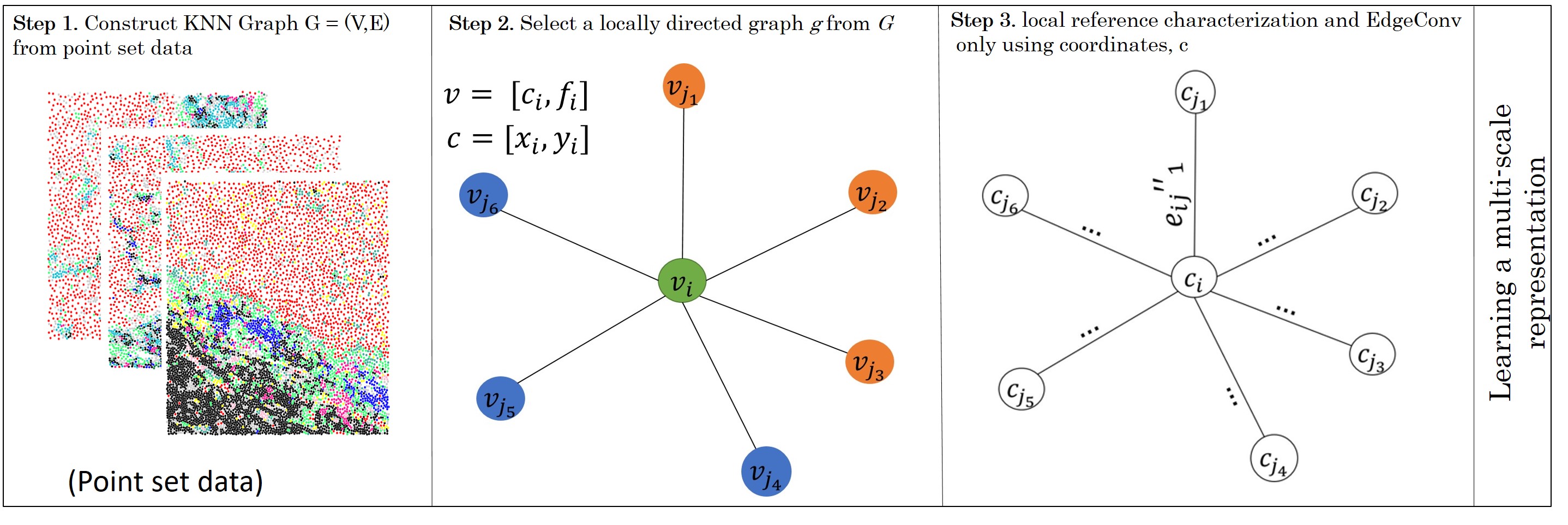}
    \caption{A multi-scale spatial representation learning using local reference characterization and EdgeConv operation. For simplicity, we are only showing one edge feature.}
    \label{PE_EdgeConv}
  \end{subfigure}
  \centering
  \begin{subfigure}[b]{1\columnwidth}
    \includegraphics[width=\linewidth, height= 4cm, width = 7.5cm]{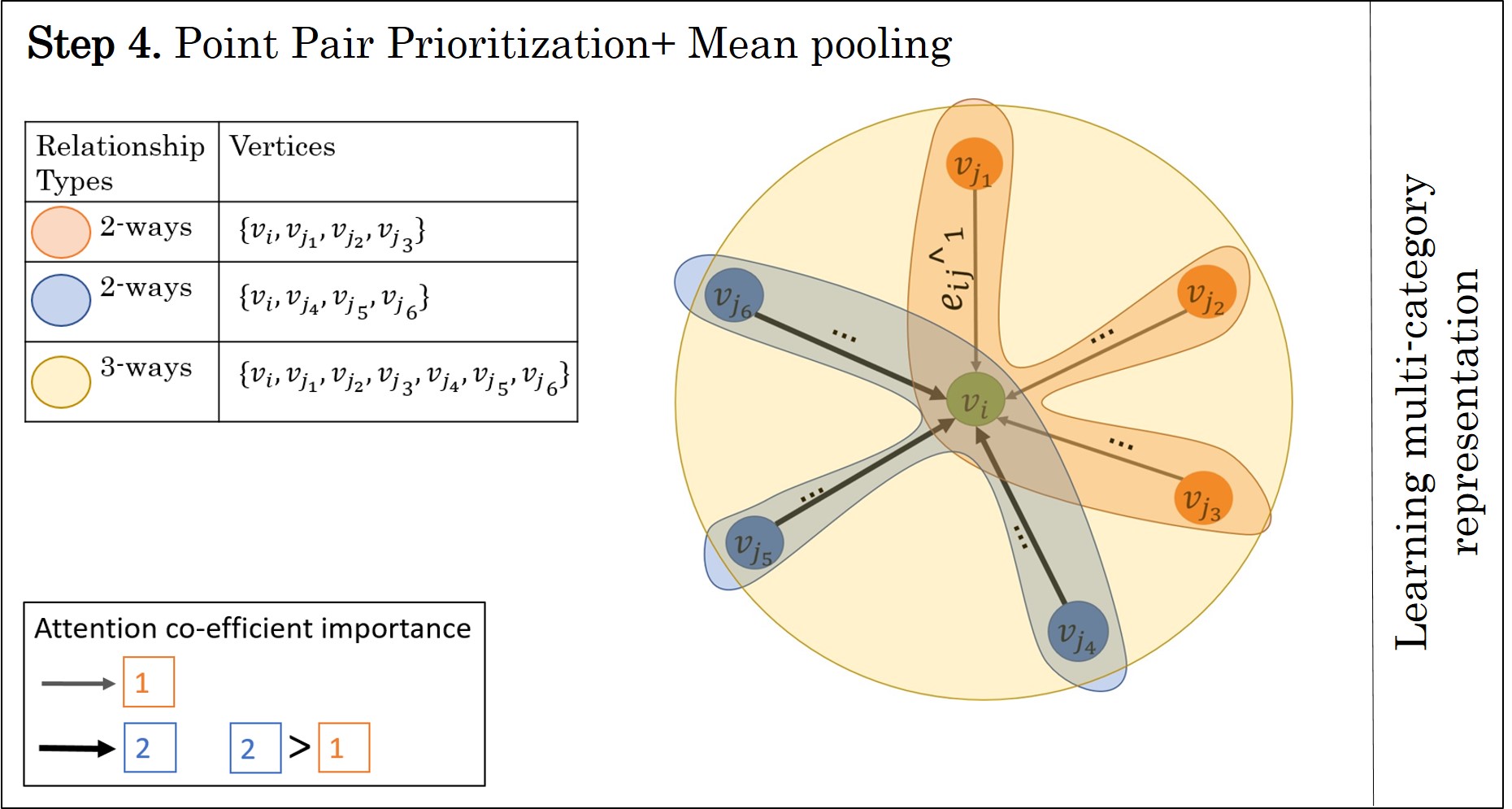}
    \caption{Learning a point pair importance $a$ based on categorical attributes (node colors) in N-way spatial relationships.}
    \vspace*{-\abovedisplayskip}
    \label{attention}
  \end{subfigure}
  \caption{The overall framework of SAMCNet.}
     \vspace*{-3\abovedisplayskip}
  \label{framework}
\end{figure}

To overcome these limitations,  we propose a \textbf{S}patial-inter-action \textbf{A}ware \textbf{M}ulti-\textbf{C}ategory deep neural \textbf{N}etwork (SAMCNet) architecture for spatial-configuration-based classification and contribute novel local reference frame characterization and point pair prioritization layers. SAMCNet provides a promising way to identify  the importance between different point pair instances and the most relevant N-way spatial relationships. As shown in Fig. \ref{PE_EdgeConv}, we first aim to provide a better way to represent spatial information using a multi-scale local reference frame characterization (LRFC) and spatial feature decomposition (described in Section \ref{representation_layer}) before applying an EdgeConv operation \cite{wang2019dynamic} . As indicated in  Fig. \ref{attention}, a  point pair prioritization sub-network is designed to specify a weight on point pair instances that are more important in an N-way spatial relationship (described in Section \ref{attention_network}). Two- and three-way spatial relationships are indicated by hyperedges connecting vertices belonging to different categories in Fig. \ref{attention}. Lastly, we use an asymmetric function (e.g., average pooling) to aggregate information across all points neighboring the center point $\hat{v_i}$. The thickness of different edges shows the contribution of distinct category pairs in the overall representation of $\hat{v_i}$. 
Fig. \ref{framework} shows the overall framework of the proposed SAMCNet.\\
\textbf{We highlight our contributions as follows:} 
\begin{itemize}
    \item We design a dynamic point pair prioritization sub-network to learn the most relevant features in N-way spatial relationships (e.g., tertiary, ternary, etc.) and use it in a Spatial-interaction Aware Multi-Category deep neural Network (SAMCNet). 
    \item We experimentally show that the proposed model outperforms existing baseline methods, and it is also computationally more efficient than the competing DNN architecture (e.g., SRNet). 
    \item We present case studies on two cancer datasets which shows the proposed model is able to identify high-order spatial patterns that are ignored by the related work, as well as the potential to advance scientific discovery.
\end{itemize}
\textbf{Scope:} We aim to identify N-way spatial relationships to help distinguish between points patterns belonging to two different classes, in which spatial relationships may vary in the strength of spatial associations. Analyzing the presence of noisy points in constructing a neighborhood graph falls outside of the scope of this paper. We also do not evaluate the effect of standard data augmentation techniques (e.g., rotation). Field trials to assess the clinical value of the proposed method also fall outside the scope of this study. Patient privacy and the propriety nature of the data prevent us from publishing the dataset.\\
\textbf{Organization:} The rest of the paper is organized as follows: Section \ref{application_domain} briefly describes an important application domain of this problem. Related work is reviewed in Section \ref{related_work}. Section \ref{problem_formulation} formally defines the problem. Section \ref{proposed_work} describes the details of our proposed work. In Section \ref{validation}, we present the experimental results and case study. 
Finally, Section \ref{conclusion_FW} concludes the paper and outlines some future research. 

\section{An Illustrative Application Domain} \label{application_domain}
The recent development of multiplex immunofluorescence (MxIF, Fig. \ref{example1}) technology has enabled exploration into the complexity of tumor-immune microenvironments within spatially-preserved metastatic tissue and in the therapeutic context of immune checkpoint inhibitors (ICI). ICI therapy works by augmenting the antitumor properties of pre-existing tumor-specific CTL, which become more efficient in infiltrating tumor masses and destroying cancer cells.  Through cyclic rounds of antibody staining, imaging and dye inactivation, MxIF technology provides a state-of-the-art method to visualize and identify many cell subtypes (e.g., immune and malignant) and their corresponding spatial coordinates using single-cell analysis of formalin-fixed paraffin-embedded (FFPE) tissue sections including metastatic melanoma lymph nodes. With continuous refinement of these techniques, it is currently possible to identify over 50 cellular phenotypic markers (e.g., CD3, FoxP3, CD14) and their corresponding cellular phenotypic and functional characteristics within a single tissue section.

\begin{figure}
    \centering
    \includegraphics[width=.7\linewidth]{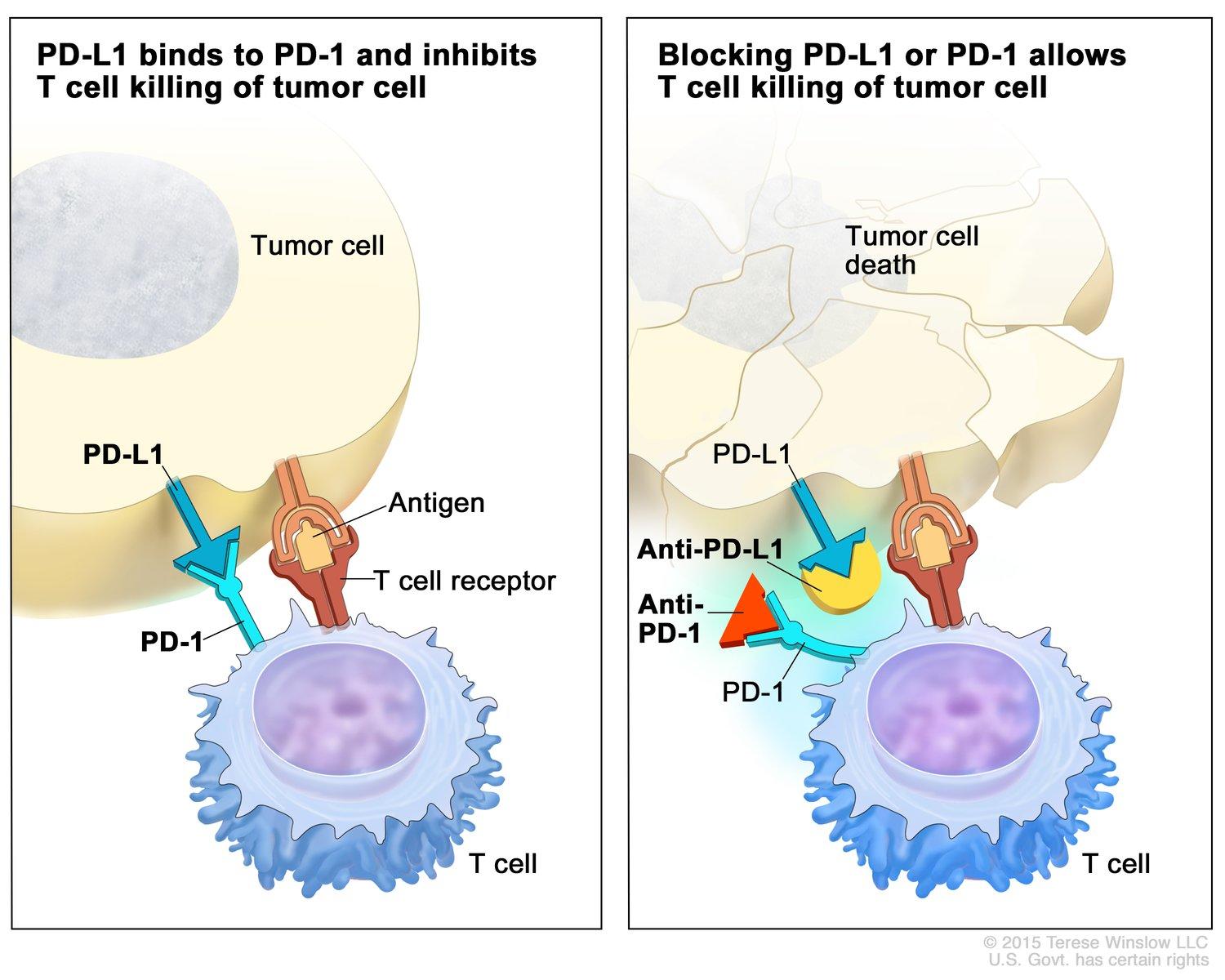}
    \caption{Importance of spatial analysis in evaluating the effectiveness of ICI therapy in killing tumor cells \cite{pdl1proximity}.}
    \label{proximity}
    \vspace*{-3\abovedisplayskip}
\end{figure}

Emerging research in this area has begun to highlight the need of an automated process to analyze the complex spatial relationships among different cellular subsets and functional states in the context of ICI therapy, which allows identifying critical intercellular interactions relevant to clinical outcomes \cite{wu2017tumor}. Furthermore, it is clinically crucial to examine the importance of cell species along with their activation states in a spatially informed manner due to the clinical implications of interactions in close spatial proximity (See Fig. \ref{proximity}).  For example, a  CTL will likely be unable to kill a cancer cell if it is nearby a tumor-associated macrophage that expresses PDL1 on its surface. In contrast, a CTL is more likely to kill a cancer cell if it is expressing Granzyme B and is not in close proximity to FoxP3-expressing regulatory T cells (Treg). We aim to provide an algorithmic description of the importance of different relationships in a tumor-microenvironment, potentially revealing insights to enhance the manual visual assessments provided by pathologists.\\

\vspace{-4\abovedisplayskip}
\section{Related Work} \label{related_work}
The related works can be classified into two major categories: (1) Data-Driven spatial quantification, (2) Machine-Constructed features using deep neural networks (DNN).\\
\textbf{Data-Driven spatial quantification:} A spatial association (i.e., spatial co-location) is an intuitive representation to help understand the spatial interactions of a multi-category point patterns by identifying a subset of points frequently located in close spatial proximity to one another \cite{shekhar2001discovering}. Spatial association interest measures (e.g., Pearson correlation, cross-k, G-cross, participation index) are commonly used in spatial data mining to quantify multi-category point patterns. Previously, a spatial association between tumor and immune cells in breast cancer digitized images of H\&E stained tissues was studied, using classical statistical methods (e.g., Pearson correlation coefficient) after imposing a spatial grid partitioning on the two-dimensional map of the cell center points \cite{maley2015ecological}. However, the classical statistics measures are sensitive to the choice of spatial partitioning. More recent work used neighbor-graph-based spatial statistical measures such as G-cross to quantify the spatial association between cancer and immune cells in lung cancer \cite{barua2018spatial}. The limitation with this approach is that it is used in isotropic space, with the same intensity regardless of measurement direction, which may not be enough to capture relevant (e.g., surrounded by) spatial interactions that might be biologically significant. 
Finally, quantifying spatial associations between different communities (e.g., at the sub-graph-level) \cite{lu2021feature, yan2019understanding} does not reveal critical information regarding the spatial relationship between distinct categorical
points within each community.\\
\textbf{Machine-Constructed features using DNN:}
A new approach that begins to address these limitations leverages machine-constructed features using a spatial-relationship aware neural network (SRNet) \cite{li2021srnet} to model spatial relationships between points of different categories. However, SRNet is limited to only binary spatial relationships, and the importance between distinct binary category pairs is assumed to be equal. Also, SRNet uses a fixed-neighborhood distance to construct the input graph to the network, and no operator is defined to work with different sized neighborhoods. All the featured-based DNN reviewed in a recent computational pathology survey \cite{duggento2021deep} use images, i.e., a regular grid, as the input. Hence, they cannot handle a simple but significantly important geometric structure, the point pattern \footnote{We provide a detailed description of DNN architectures for point cloud data \cite{wang2019dynamic, qi2017pointnet, qi2017pointnet++} in appendix Section \ref{other_related_works}; these DNN architectures are limited in learning spatial relationships in multi-categorical point patterns.}.  
\section{Problem Formulation} \label{problem_formulation}
\subsection{Problem Statement} The main goal of this study is to build a spatial-configuration-based classifier to distinguish between multi-category point patterns that belong to  two different classes. The primary objective is to achieve a high solution quality (e.g., accuracy). In addition, we aim to identify the most relevant N-way spatial relationships that help distinguish between different classes. Fig. \ref{example1} shows an example of the point patterns of two fields of view of MxIF images at the tumor-margin that need to be classified into two unique classes, namely "responder" and "non-responder", reflecting different clinical outcomes.



There are three key building blocks in constructing our model. The first is a multi-scale local reference frame characterization (LRFC), which takes as input a neighborhood graph that models the spatial distribution of one point and its neighbors. The second block uses an EdgeConv \cite{wang2019dynamic} to learn local and global information and a semantic representation by dynamically updating the neighborhood graph at each layer (see Section \ref{DGCNN}). The third is a prioritization sub-network to distinguish between point pair instances that belong to different categories by learning the importance of each distinctive pair, we use an asymmetric function (e.g., average pool) to aggregate the information of one point and all its neighbors (see Section \ref{attention_network}).
\vspace*{-\abovedisplayskip}
\section{The proposed work (SAMCNet)} \label{proposed_work}
The primary objective of the proposed neural network architecture is to learn N-way spatial relationships in a multi-category point pattern. The main difference between SAMCNet and traditional data-driven association interest measures is LRFC, which allows categorical points belonging to different distributions (e.g., clustering versus even distribution) to be represented through a multi-scale representation, overcoming the inefficiency of intrinsically single-scale methods like radial basis function kernels or discretization. Furthermore, SAMCNet differs from the competing DNN architecture SRNet in two distinct ways. First, it incorporates a point pair prioritization sub-network, which learns the importance of point pairs in N-way spatial relationships based on their categorical attributes. Second, the connectivity of nodes in a locally connected graph allows SAMCNet to learn relevant high-order spatial patterns based on the input $k$ nearest neighbors and aggregation choice. 


\begin{figure*}
    \centering
    \includegraphics[width=\linewidth, height = 7.3cm]{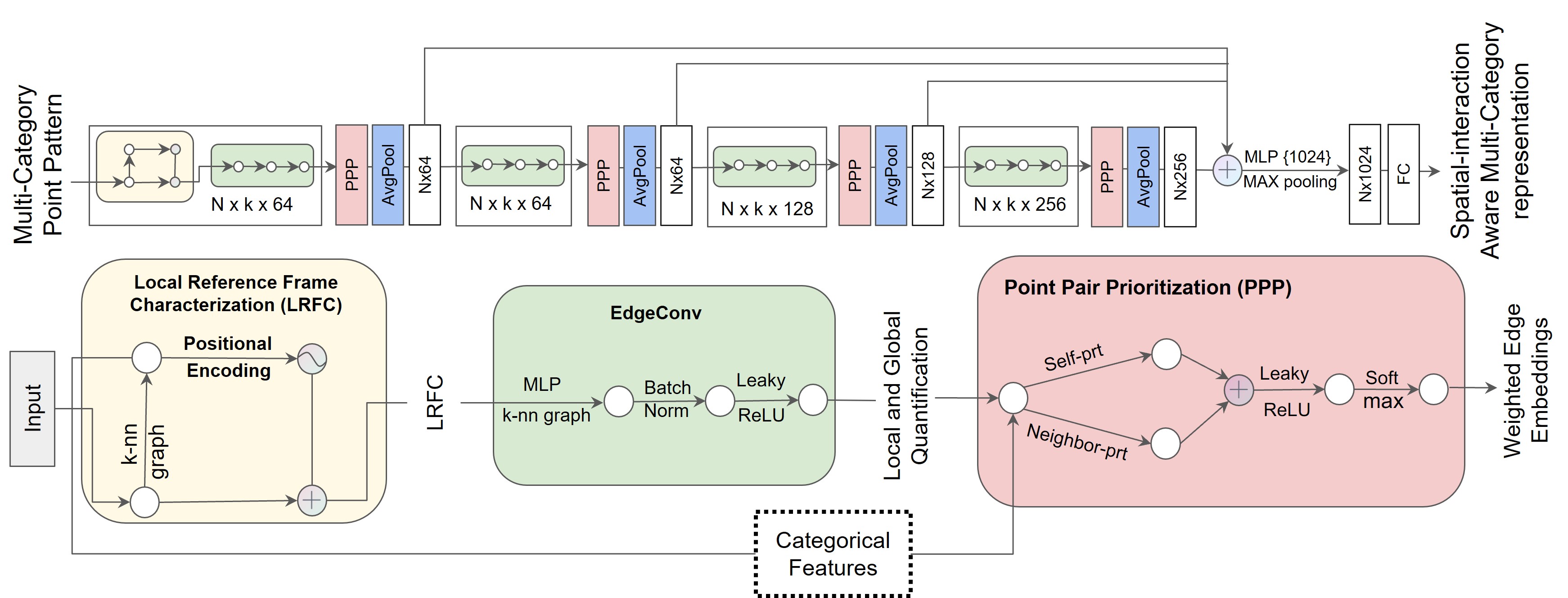}
    \caption{\textbf{SAMCNet Architecture.} The network architecture takes as input a multi-category point set containing n points, where a local reference frame characterization (LRFC) layer calculates an embedding for each point using its coordinates and neighborhood spatial distribution. Embeddings are then passed into the EdgeConv layer to specify an edge feature set of size k for each point. The categorical features are passed using a skip connection, where a point pair prioritization sub-network calculates the importance between a point and its k nearest neighbors belonging to different category pairs. Lastly, an average pooling aggregates information from all $k$ points to the center point.  Notice that since a K-nn graph is built in the LRFC layer, the reconstructing of the K-nn graph in the first EdgeConv layer is skipped.}
    \label{architecture}
\end{figure*}

\subsection{Local Reference Frame Characterization} \label{representation_layer}
Given a point set $P= \{p_i = (c_i,f_i) | p_1, ... , p_n \}$, where $c_i=(x_i, y_i)$ are the spatial coordinates and $f_i$ is the categorical attribute \footnote{The categorical feature $f_i$ associated with each $p_i$ is preserved throughout the network using a skip connection for point pair prioritization sub-network computations (See Fig. \ref{architecture}).}, we compute directed graph $G=(V, E)$, where $V$ and $E$ are the vertices and edges. We construct $G$ as the $k$-nearest neighbor graph of each $c_i$ point  $\in {\rm I\!R}^F$, where $E=V\times K$. Note that the neighborhood graph of each consecutive layer in SAMCNet relies on the output of the preceding layer, which is dynamically updated based on dimension $F$, which represents the feature dimensionality of the given layer. For example, in the beginning, the neighborhood graph $G$ is constructed as the k-nearest neighbor points of each $c_i$ point  $\in {\rm I\!R}^2$, which represent the spatial coordinates. This approach allows the network to learn how to build the graph G utilized in each layer, rather than using a fixed constant graph established before the network is evaluated. Reconstructing the neighborhood graph in the embedding space produced by the hidden layer using nearest neighbors is empirically beneficial in related classification tasks, as shown in \cite{wang2019dynamic}.

The next step is to use local reference frame characterization (LRFC) to model the distribution of one point and its neighbor by only using spatial coordinates. This technique allows us to model the relative distance between a given point $c_i$ with the respect to its nearest points $c_{j}$, where $1\leq j \leq k$, into a corresponding edge $e_{ij}'$. The intuition behind the LRFC is that spatial coordinates are illustrative location indicators; using discretization or feed-forward neural network techniques is insufficient to capture the spatial distribution due to the lack of feature decomposition between spatial and categorical attributes. 
Inspired by a multi-scale periodic representation of grid cells in mammals \cite{abbott2014nobel} and a vector representation of self-position \cite{gao2018learning}, Mai et al. \cite{mai2020multi} proposed a multi-scale embedding, namely positional encoding $PE$, which uses sine and cosine functions of different frequencies to present positions in space. We adopt this idea in our network as follows. 

Given a point $c_i$ in a studied 2D space, $e[c_i] =$ $EdgeConv-(PE_s(c_i))$, where $PE(c_i)$ is a multi-scale representation $s_j$, $1 \leq j \leq s$, to capture the distribution of mixture multi-category point patterns. The overall formulation of local reference frame characterization (LRFC) is as follows:
\begin{equation}\label{scaling}
\vspace{-0.5cm}
    PE_s(c_i) = [PE_{s,1}(c_i); ...; PE_{s,s}(c_i)],
\end{equation}

\begin{equation}\label{cos_sin}
    \begin{split}
        PE_{s,j}(c_i) = [\cos(\frac{\langle c_i, a_j \rangle}{\lambda_{min} \cdot g^{s/(S-1)}}); & \sin(\frac{\langle c_i, a_j \rangle}{\lambda_{min} \cdot g^{s/(S-1)}})], \\
        & \forall j = 1, 2, 3,
    \end{split}
\end{equation}
where $a_1 = [1, 0]^T, a_2=[-1/2, \sqrt{3}/2]^T, a_3=[-1/2, -\sqrt{3}/2]^T$ are unit vectors, the angles between every pair of vectors is $2\pi/3$, $\lambda_{min},\lambda_{max}$ are the minimum and maximum grid scales, and $g= \frac{\lambda{max}}{\lambda_{min}}$. We define the input to PE as the distance between the center point $c_i$ and its k-nearest neighbors $c_j$ as $|PE(c_i) - PE(c_j)|)$, where $1 \leq j \leq k$.  

\subsection{Local and Global Quantification} \label{DGCNN}


The EdgeConv operation is defined as edge feature $e_{ij} = h_{\Theta} (c_i,c_j)$, where $h_{\Theta}$ : $R^F \times R^F$ $\rightarrow$ $R^{F^{\prime}}$  is a nonlinear function with a set of learnable parameters $\Theta$. Lastly, an asymmetric operation (e.g., $\sum$ or Max) is applied to aggregate information along all the edge features neighboring center node $c_i$. The choice of $h_{\Theta}$ is critical in defining EdgeConv, such as using the dot product between a set of filters $\Theta = \{\theta_1, ..., \theta_M \}$ and image pixels $x_j$ in a regular grid and aggregating information using $\sum$ results in a standard convolution. A detailed discussion of different forms of $h_\Theta$ can be found in \cite{wang2019dynamic}. 

We have adapted the EdgeConv operation from DGCNN \cite{wang2019dynamic} in our network to learn both global shape structure, captured by the center coordinates $c_i$, and local neighborhood information, captured by $|c_i - c_j|$. The overall formulation is as follows:
\vspace*{-\abovedisplayskip}
\begin{equation}
    e^{''}_{ij} = leakyrelu( \theta_m . |PE(c_i) - PE(c_j)| + \phi_m . c_i), 
\end{equation}
where $\theta_m$ and $\phi_m$ are learnable parameter for local and global information, respectively. $PE$ is the positional embedding to represent relative distances along each edge starting at $c_i$.

\subsection{ Point Pair Prioritization  Sub-Network} \label{attention_network}
Thus far, we have built the graph and defined the edge embeddings in terms of strictly spatial features. If we follow existing point patterns graph-based  DNN architectures (e.g., Pointnet++ \cite{qi2017pointnet++}, DGCNN \cite{wang2019dynamic}), we would simply concatenate the categorical features into the embedded feature space.  However, the importance of interactions between vertices of categorical features $f_i$ and $f_{j\in \mathcal{N}_i}$ would not be learned in this way. As a result, the model would be confined to learning individual category features. Instead, the classifier should learn how to correctly weight diverse point pair associations as a stronger inductive bias. To this end, we propose a  point pair prioritization layer to learn the importance (i.e., strength) of the spatial relationship between different category pairs, followed by an average pooling layer to weigh different subsets accordingly. As a whole, this layer is analogous to a weighted average pooling function, where the weights correspond to the importance of the categorical interaction.

The input to this layer is an edge embedding $e_{ij}''$, which is the output from the EdgeConv layer. In the prioritization  layer, we first derive $\hat{e}_{ij}$, an edge embedding augmented by the strength of categorical pairwise association:
\vspace*{-\abovedisplayskip}
\begin{equation}
    \hat{e}_{ij} = \vec{a}_{f_if_j}^TWe_{ij}'',
    \vspace*{-\abovedisplayskip}
    \label{importance}
\end{equation} 
where $W$ is a learnable linear transformation on the original embedding to aid prioritization expressivity, and $\vec{a}_{f_if_j}$ is our learned pairwise association weight vector for categorical point pair features $(f_i, f_j)$. 

In this formulation, we have included $f_{j\in \mathcal{N}_i}$, where $\vec{a}_{f_if_j}$ is a learned self-weighting based only on the categorical feature of $v_i$. We also note that interactions are assumed invariant with respect to the ordering of the categories; for example, $\alpha_{\mathcal{C}_1\mathcal{C}_2}	\equiv\alpha_{\mathcal{C}_2\mathcal{C}_1}$. Similar to other prioritization (i.e., attention) layers, we then apply a LeakyReLU ($\text{LR}$) activation, followed by a softmax function, resulting in the normalized pairwise association:
\vspace*{-\abovedisplayskip}
\begin{equation}
    \vspace*{-\abovedisplayskip}
    \alpha_{f_if_j} = \frac{\exp(\text{LR}(\hat{e}_{ij}))}{ \sum_{k \in \mathcal{N}_i} \exp(\text{LR}(\hat{e}_{ij}))},
\end{equation}\\
%
where $\alpha_{f_if_j}$  is the learned categorical pairwise association for each neighbor, such that $ f_j \in {\rm I\!R}^k $. With this normalized attention coefficient, $\alpha_{f_if_j}$, we can calculate the weighted average pooling and produce the final vertex embedding:
\vspace*{-\abovedisplayskip}
\begin{equation}
\vspace*{-\abovedisplayskip}
\label{eq:vertex_relationship_embedding}
    \hat{v_i}=\sigma\left(\frac{1}{|\mathcal{N}_i|}\sum_{j\in \mathcal{N}_i} \alpha_{f_if_j} W e_{ij}''\right)
\end{equation}

This formulation can be extended to $K$ heads, following other prioritization networks such as GAT \cite{velivckovic2017graph}, where each head learns a separate categorical pairwise association $\alpha_{f_if_j}^k$ and linear transformation weight $W^k$ followed by an aggregation operation (AGG) over the different head outputs:
\begin{equation}
\label{eq:vertex_relationship_embedding}
    \hat{v_i}=\text{AGG}_{k=1}^K \sigma\left(\frac{1}{|\mathcal{N}_i|}\sum_{j\in \mathcal{N}_i} \alpha_{f_if_j}^k W^k e_{ij}''\right),
    \vspace*{-\abovedisplayskip}
\end{equation}
where $\sigma$ is a non-linear activation function such as LeakyRelu and the aggregation operation can take the form of an average or a concatenation. Since this layer preserves the identity of the center vertex, it can also be extended to multiple layers of the network by maintaining the categorical features of vertices between layers with a skip connection. We do this by adding  point pair prioritization  after each layer's EdgeConv operation. In the context of hierarchical feature learning, our network is therefore effectively capable of learning the importance of categorical N-way interactions in a hierarchical feature space.

Finally, we note that the choice of aggregation is not limited to average pooling; for example, one may choose to select a large number of k-nearest neighbors when building the graph, while taking only a top-$k'$ subset of highest features to pool, in order to filter out an overpowering number of weak interactions. Max pooling can be demonstrated as a special case of this concept, where only the top-1 of a neighbor's features is selected.
\begin{table*}\scriptsize
\caption{Model performance on all datasets.}
\vspace*{-2\abovedisplayskip}
\hspace*{-1cm}
\begin{tabular}{|l|l|l|l|l|} \hline
Model & \multicolumn{4}{c|}{Tumor-Margin}\\ \hline
& Precision & Recall & F1-Score & Acc\\ \hline
PI + DT & 0.93 & 0.93   & 0.93& 0.93 \\ \hline
Cross-K + DT & 0.60 & 0.50   & 0.44& 0.50 \\ \hline
PI + RF & 0.85 & 0.82   & 0.82& 0.82 \\ \hline
Cross-K + RF & 0.61 & 0.61   & 0.61& 0.60 \\ \hline
PI + NN & 0.32 & 0.57   & 0.41& 0.57 \\ \hline
Cross-K + NN & 0.71 & 0.70   & 0.71& 0.70 \\ \hline
PointNet   & 0.82 & 0.68  & 0.66 & 0.68 \\ \hline
DGCNN   & 0.76 & 0.46   & 0.33 & 0.46 \\ \hline
SRNet  & 0.90  & 0.89  & 0.89 & 0.89 \\ \hline
\begin{tabular}[c]{@{}l@{}} \textbf{SAMCNet}  \\ \textbf{(ours)} \end{tabular}  & \textbf{0.97} & \textbf{0.96}   & \textbf{0.96} & \textbf{0.96} \\ \hline
\end{tabular}
\hfill
\begin{tabular}{|l|l|l|l|l|} \hline
Model & \multicolumn{4}{c|}{Tumor-Core}\\ \hline
& Precision & Recall & F1-Score & Acc\\ \hline
PI + DT & 0.70  & 0.62   & 0.64 & 0.62 \\ 
\hline
Cross-K + DT & 0.52 & 0.72   & 0.60 & 0.72 \\ 
\hline
PI + RF & 0.73 & 0.75   & 0.70 & 0.75 \\ 
\hline
Cross-K + RF & 0.52 & 0.72   & 0.60 & 0.72 \\ 
\hline
PI + NN & 0.75 & 0.75   & 0.75 & 0.75 \\ 
\hline
Cross-K + NN & 0.52 & 0.72   & 0.60 & 0.72 \\ 
\hline
PointNet   & 0.68 & 0.67   & 0.67 & 0.67 \\ \hline
DGCNN    & 0.49 & 0.33 & 0.29 & 0.33 \\ \hline
SRNet   & 0.96 & 0.95  & 0.95 & 0.95 \\ 
\hline
\begin{tabular}[c]{@{}l@{}} \textbf{SAMCNet}  \\\textbf{(ours)} \end{tabular} & \textbf{0.96}  & \textbf{0.95} & \textbf{0.95} & \textbf{0.95}  \\
\hline
\end{tabular}
\hfill
\begin{tabular}{|l|l|l|l|l|} \hline
Model & \multicolumn{4}{c|}{Disease}\\ \hline
& Precision & Recall & F1-Score & Acc\\ \hline
PI + DT & 0.76 & 0.77   & 0.76 & 0.77 \\ \hline
Cross-K + DT & 0.85 & 0.85   & 0.85 & 0.85 \\ \hline
PI + RF & 0.82 & 0.80   & 0.78 & 0.80 \\ \hline
Cross-K + RF & 0.84 & 0.82 & 0.80 & 0.82 \\ \hline
PI + NN & 0.84 & 0.78   & 0.75 & 0.78 \\ \hline
Cross-K + NN & 0.40 & 0.63   & 0.49 & 0.63 \\ \hline
PointNet   & 0.76 & 0.6 & 0.48 & 0.60 \\ \hline
DGCNN   & 0.54 & 0.57   & 0.46 & 0.57 \\ \hline
SRNet   & 0.78 & 0.77  & 0.76 & 0.77  \\ \hline
\begin{tabular}[c]{@{}l@{}} \textbf{SAMCNet}  \\ \textbf{(ours)} \end{tabular}  & \textbf{0.94} & \textbf{0.94}   & \textbf{0.94} & \textbf{0.94} \\ \hline
\end{tabular}
\hspace*{-0.32cm}
\label{compare_analysis}
\vspace*{-\abovedisplayskip}
\end{table*}

\section{Validation} \label{validation}
\subsection{Experimental Settings}
\textbf{Evaluation Tasks:} We validated our proposed approach with (1) a \textbf{\textit{comparative analysis}} to evaluate the proposed SAMCNet against classical spatial association intereset measures and state-of-the-art DNN architectures on this problem, (2) a \textbf{\textit{sensitivity analysis}} to evaluate the impact of key building blocks (e.g., self-prioritization, neighboring prioritization, etc.) and with key parameters on selected performance metrics (See appendix Section \ref{more_sensitivity}), and (3) a feature selection analysis to evaluate the \textbf{impact of prioritization sub-network} to learn the importance of different point pair instances and identify \textbf{the most relevant N-way spatial relationships}.\\   
\textbf{Model Architecture:}
Fig. \ref{architecture} shows the network architecture. The proposed SAMCNet was implemented in Pytorch. For local reference frame characterization, the grid-scale, minimal grid cell size and maximal grid cell size were set to 5, 1, and 100, respectively. The number of k nearest neighbors was set to 6. We followed the same settings for the 4 EdgeConv layers, residual block connection, batch normalization, activation functions, and dropout as described in \cite{wang2019dynamic}. We used the Adam optimization algorithm with a learning rate of $10^{-3}$ and cross-entropy loss for 200 epochs to train the SAMCNet. The batch size, momentum, and prioritization heads were set to 7, 1, 0.9, respectively. All hyper-parameters were set through tuning on the validation set. DNN candidate methods were tested with the same setting described above.\\
\textbf{Baseline Methods \footnote{Traditional classifiers and fully connected neural network were implemented with the Python scikit-learn package \cite{scikit-learn}, with hyper-parameters set to default values unless otherwise stated. }:} We compared our proposed framework on selected classification metrics with the following baseline methods. Point patterns composed of hand-constructed features using participation index (PI) \cite{shekhar2001discovering}, then spatial association interest measure values were fed into a \textbf{(1)} decision tree with a max depth of 2 \textbf{(PI+DT)}, \textbf{(2)} Random Forest with similar depth \textbf{(PI + RF)}, and \textbf{(3)} fully connected neural network \textbf{(PI + NN)} with four Relu hidden layers and 2048 neurons. In similar settings, we used point patterns composed of cross-k \cite{intro_SDM} values fed into similar classifiers previously described, giving us three more candidate methods: \textbf{(4) cross-k + DT}, \textbf{(5) cross-k + RF}, and \textbf{(6) cross-k + NN}. We used a fixed-neighborhood distance of 50 pixels to construct participation index and cross-k values. We also evaluated proposed model with state-of-the-art DNN architectures: \textbf{(7) PointNet} \cite{qi2017pointnet++},  a neural network architectures that directly consume point sets for applications ranging from object classification to part segmentation; \textbf{(8) DGCNN} \cite{wang2019dynamic}, a dynamic graph convolutional neural network architecture for CNN-based high-level point cloud tasks such as classification and segmentation; \textbf{(9) SRNet} \cite{li2021srnet}, a DNN architecture for binary spatial relationships in multi-category point patterns.\\
\textbf{Dataset:} \label{dataset} \label{dataset_desc} 
Experiments were conducted on two multi-category point pattern cancer datasets from MxIF images. The first dataset was used for two distinct classification tasks,  (1) \textbf{tumor-margin} classification and (2) \textbf{tumor-core} classification. The second dataset was used for a (3) \textbf{disease} classification task. In the \textbf{tumor-margin} classification task, we used 145 FOV point sets indicating two different clinical outcomes of ICI therapy, 68 of which were labeled as responders, and 77 labeled as non-responders for individual who progressed and experienced recurrence in less than a year. We used 103 FOVs point sets in the \textbf{tumor-core} classification task, 30 of which were labeled as responders and 73 non-responders, extracted from tumor area of metastatic lymph nodes. In the \textbf{disease} classification task, we used 143 point sets of chronic pancreatitis and  53 pancreatic ductal adenocarcinomas (PDAC).\\ 
\textbf{Evaluation Metrics:} The model performance was measured by using the weighted average of precision, recall, F1-score, and accuracy (ACC).\\
\textbf{Data Preparation:} In each classification task, we divided the data into 80\% training and 20\% testing. Ten percent of the training set was selected to be the validation set. Due to the limited number of learning samples, we used data augmentation techniques, whereby each learning sample was rotated 12 degrees clockwise five times during the training procedure. We restricted rotation to only five times due to potential overfitting issues. We uniformly sampled 1,024 points from each point set for the underlying classification task.\newline
\textbf{Platform:} We used K40 GPU composed of 40 Haswell Xeon E5-2680 v3 nodes. Each node has 128 GB of RAM and 2 NVidia Tesla K40m GPUs. Each K40m GPU has 11 GB of RAM and 2880 CUDA cores.
\subsection{Experimental Results} 
\textbf{Comparative Analysis:} We tested the candidate methods on tree classification tasks described in Section \ref{dataset_desc}. Results on selected classification metrics are presented in Table \ref{compare_analysis}. Results show the superiority of the proposed SAMCNet over traditional data-driven spatial association interest measures and existing DNN competition (i.e., PointNet, DGCNN, SRNet). 
Most notably, we were able to improve accuracy over SRNet by a margin of 7.0\%, and 17.0\% on tumor-margin, and disease classifications, respectively. These results suggest that local reference frame characterization (LRFC) and specifying different weights to points of the same neighborhood with the distinct categorical attribute are beneficial.
\begin{table}\scriptsize
\centering
\caption{Model time complexity on all datasets.}
\vspace*{-2\abovedisplayskip}
\label{time_complexity}
\begin{tabular}{|c|c|c|c|} 
\hline
\multirow{2}{*}{\begin{tabular}[c]{@{}c@{}}\\Model\end{tabular}} & \multicolumn{3}{c|}{Time (second)}       \\ 
\cline{2-4}& tumor-margin & tumor-core & disease  \\ 
\hline
SRNet    & 30.51          & 25.34       & 19.18      \\ 
\hline
SAMCNet & 4.04           & 2.41         & 5.77       \\
\hline
\end{tabular}
\vspace*{-3\abovedisplayskip}
\end{table}

In addition, we used the model inference time on three classification tasks as a measure of the model's computational time complexity and examined the trade-off between time complexity and classification accuracy. In this experiment, we only compared our proposed model with SRNet as a direct competitor, which is specifically designed to learn spatial associations in multi-category point patterns. SAMCNet is not only more accurate than SRNet, but it also runs faster from 3 to 10 times faster on the three classification tasks. Table \ref{time_complexity} provides the details on each classification task. Pytorch Profiler \footnote{https://pytorch.org/tutorials/recipes/recipes/profiler_recipe.html} was used to measure the inference time across different candidate methods.\\
\textbf{Sensitivity Analysis:} To evaluate the performance of the primary building pieces inside our suggested DNN architecture, we asked one question: How does the model perform in the presence and absence of important components? To answer this, we incrementally added key elements of the model (e.g., using only local reference frame characterization (LRFC), only using self-prioritization (i.e., self-prt), and only using neighbor-prioritization (i.e., neighbor-prt) , etc.) and assessed performance using a variety of classification measures. 

Results on selected classification metrics are presented in Table \ref{sensitivity_analysis}. The results show that using a prioritization sub-network is beneficial to filter out the exponential number of weak interactions caused by center points (e.g., center cells) and neighboring points (e.g., neighboring cells). In addition, it can be observed that the local reference frame characterization (LRFC) plays a critical role in representing relative distances and mixture distribution caused by neighboring points (e.g., cells), where LRFC combined with a prioritization sub-network (i.e., self or neighbor prioritization) provides better classification performance in most cases. This last result suggests that the proposed model performs the best when the local reference frame characterization layer is integrated with the point pair prioritization sub-network as a whole.\\ 
\textbf{Impact of Prioritization Sub-network}:
The goal of this experiment was to demonstrate the interpretability of SAMCNet by measuring the impact of various point pair values (e.g., contribution between distinct cells) for distinguishing between point patterns of different classes (e.g., responder and non-responder). However, non-linear activation functions in DNN architectures are required for learning complex configurations, making ML models hard to interpret and this task challenging. To address this problem, we separated the feature vector indicating the relevance of distinct point pairs from the pooling layer and non-linear activation functions . We transformed the feature vector from each distinct point pair to a scaler value $\widetilde{pp_i}$ using a vector norm \cite{kobayashi2020attention} to measure the magnitude (i.e., importance) of the learned associations. We divided each $\widetilde{pp_i}$ by the maximum value found across all point pair scalers to further normalize them for a direct comparison.
\begin{table*}\scriptsize
\caption{Model performance based on different building blocks on all datasets.}
\vspace*{-2\abovedisplayskip}
\hspace*{-0.25cm}
\begin{tabular}{|l|l|l|l|l|} \hline
Model & \multicolumn{4}{c|}{Tumor-Margin}\\ \hline
& Precision & Recall & F1-Score & Acc\\ \hline
Only LRFC & 0.64 & 0.61 & 0.6 & 0.61 \\ 
\hline
Only self-prt & 0.94 & 0.93 & 0.93 & 0.93 \\ 
\hline
Neighbor-prt & 0.86 & 0.86 & 0.85 & 0.86 \\ 
\hline
\begin{tabular}[c]{@{}l@{}}LRFC+ \\self-prt\end{tabular}  & 0.89 & 0.89 & 0.89 & 0.89 \\ 
\hline
\begin{tabular}[c]{@{}l@{}}LRFC+ \\Neighbor-prt\end{tabular} & 0.93 & 0.93 & 0.93 & 0.93 \\ 
\hline
\begin{tabular}[c]{@{}l@{}}self-prt + \\Neighbor-prt\end{tabular}  & 0.86 & 0.82 & 0.81 & 0.82 \\ 
\hline
Entire model & 0.97 & 0.96 & 0.96 & 0.96 \\
\hline
\end{tabular}
\hfill
\begin{tabular}{|l|l|l|l|l|} \hline
Model & \multicolumn{4}{c|}{Tumor-Core}\\ \hline
& Precision & Recall & F1-Score & Acc\\ \hline
Only LRFC & 0.44 & 0.66 & 0.53 & 0.67 \\ 
\hline
Only self-prt & 0.83 & 0.81 & 0.81 & 0.81 \\ 
\hline
Neighbor-prt & 0.83 & 0.81 & 0.81 & 0.81 \\ 
\hline
\begin{tabular}[c]{@{}l@{}}LRFC+ \\self-prt\end{tabular}  & 0.87 & 0.86 & 0.86 & 0.86 \\ 
\hline
\begin{tabular}[c]{@{}l@{}}LRFC+ \\Neighbor-prt\end{tabular}  & 0.87 & 0.86 & 0.86 & 0.86 \\ 
\hline
\begin{tabular}[c]{@{}l@{}}self-prt + \\Neighbor-prt\end{tabular}  & 0.85 & 0.81 & 0.78 & 0.81 \\ 
\hline
Entire model & 0.96 & 0.95 & 0.95 & 0.95 \\
\hline
\end{tabular}
\hfill
\begin{tabular}{|l|l|l|l|l|} \hline
Model & \multicolumn{4}{c|}{Disease}\\ \hline
& Precision & Recall & F1-Score & Acc\\ \hline
Only LRFC & 0.66 & 0.66 & 0.66 & 0.66 \\ 
\hline
Only self-prt & 0.84 & 0.83 & 0.82 & 0.83 \\ 
\hline
Neighbor-prt & 0.83 & 0.83 & 0.83 & 0.83 \\ 
\hline
\begin{tabular}[c]{@{}l@{}}LRFC+ \\self-prt\end{tabular} & 0.87 & 0.83 & 0.82 & 0.83 \\ 
\hline
\begin{tabular}[c]{@{}l@{}}LRFC+ \\Neighbor-prt\end{tabular} & 0.9 & 0.89 & 0.88 & 0.89 \\ 
\hline
\begin{tabular}[c]{@{}l@{}}self-prt + \\Neighbor-prt\end{tabular}  & 0.87 & 0.86 & 0.86 & 0.86 \\ 
\hline
Entire model & 0.94 & 0.94 & 0.94 & 0.94 \\
\hline
\end{tabular}
\hspace*{-0.25cm}
\label{sensitivity_analysis}
\end{table*}

The feature vectors were composed of different distinct cell values extracted at layer-1 and layer-4 from the point pair prioritization sub-network, an indication of the ability of SAMCNet to learn hierarchical feature representations. These values present vector norm of $\vec{a}_{f_if_j}^T$ in equation \ref{importance}. Results are presented in Fig. \ref{importance}, where it can be observed that the spatial interactions between points within the same category (e.g., \{Tumor Cell, Tumor Cell\}, \{Vasculature, Vasculature\}) remain critical across all layers. By contrast,  the spatial interactions between distinct pairs (e.g., \{Tumor Cell, Macrophage\}) are adjusted through different prioritization layers, but they remain relatively important in the learning N-way spatial relationships separating two distinct classes (e.g., responder and non-responder) in the tumor-core. Note that we chose the tumor enriched areas of the lymph node (tumor-core) to interpret the proposed SAMCNet because it is primarily used in oncology analysis to determine the efficacy of ICI therapy by examining the various spatial interactions and variability between different cell species.\\
\textbf{Most Relevant N-way Spatial Relationships:}
Thus far, we have demonstrated the most distinctive point pairs; but the main idea of this work is to show SAMCNet's ability to identify the most significant high-order spatial interactions. Hence, each sample point pattern was represented by extracting its corresponding feature vectors composed of all N-way spatial relationships (e.g., tertiary, ternary) with respect to its center and neighboring points (e.g., different cell categories). To be more precise, the trained model was used to extract features after the point pair prioritization network at layer-4, where the model has learned both spatial and categorical associations. Thus, we have a feature vector $\vec{v} = Mean([k'' * f'])$ for each N-way spatial relationship,  where it is composed of an aggregation (e.g., mean) over $k''$ as point categories located around the center point and $f'$ is the embedded feature space (e.g., 256). For example, having a tumor cell as a center cell (i.e., data point) and unique counts of macrophages and neutrophils as neighboring cells is an instance of a 3-way spatial association.   

We evaluated the importance of the identified spatial relationships, namely, SAMCNet representation for different category subsets, through permutation feature importance. This metric measures the importance of a given feature by the increase observed in the prediction error caused by randomly shuffling the feature space. The relevance of the discovered N-way spatial relationships was tested in this experiment by the classification accuracy after exchanging the corresponding elements in the representation vectors. The top five most relevant spatial associations found within the tumor-core are shown in Table \ref{rels_importance}.  

\begin{table}\scriptsize
\centering
\caption{Most relevant spatial relationships}
\vspace*{-2\abovedisplayskip}
\begin{tabular}{|l|l|l|} 
\hline
Rank & Center cell     & Neighboring cells                       \\ 
\hline
1    & {[}Macrophage]  & {[}Macrophage, Neutrophil, Tumor Cell]  \\ 
\hline
2    & {[}Tumor Cell]  & {[}Tumor Cell, Vasculature]             \\ 
\hline
3    & {[}Tumor Cell]  & {[}Macrophage, Tumor Cell]              \\ 
\hline
4    & {[}Vasculature] & {[}Tumor Cell, Vasculature]             \\ 
\hline
5    & {[}Vasculature] & {[}Neutrophil, Tumor Cell,Vasculature]  \\
\hline
\end{tabular}
\label{rels_importance}
\vspace*{-3\abovedisplayskip}
\end{table}

\subsection{Clinical Implications (Interpretation of results)}
The pro-tumor (non-responder) relationship between tumor-associated macrophages and tumor-associated neutrophils has been studied in past works \cite{chandra2021colorectal}, although its nature remains not entirely clear. Tumor-associated macrophages appear to play a protective role against antitumor immunotherapy, and the association of tumor-associated macrophages with tumor cells preferentially in patients who did not respond is consistent with established biology. However, added to this is the effect of tumor-associated neutrophils which also appear to promote tumor cell survival and lack of response to immunotherapy. The latter is less well understood, with notable findings indicating that tumor-associated neutrophils are relevant to tumor progression but not necessarily to immunotherapy resistance or a relationship with tumor-associated macrophages. 

These findings suggest a previously unknown shared biology among these two populations of myeloid cells (macrophages and neutrophils) and provide new insight into the possibility of a relationship between tumor-associated macrophages and tumor-associated neutrophils as they engage tumor cells in the tumor core. This is an intriguing pattern that has not yet been studied. Future research is warranted to understand the relationships of macrophages, tumor cells, and neutrophils sub-population in these interactions. 

\begin{figure}
    \centering
    \includegraphics[width = 1.05\linewidth, height = 5.5cm]{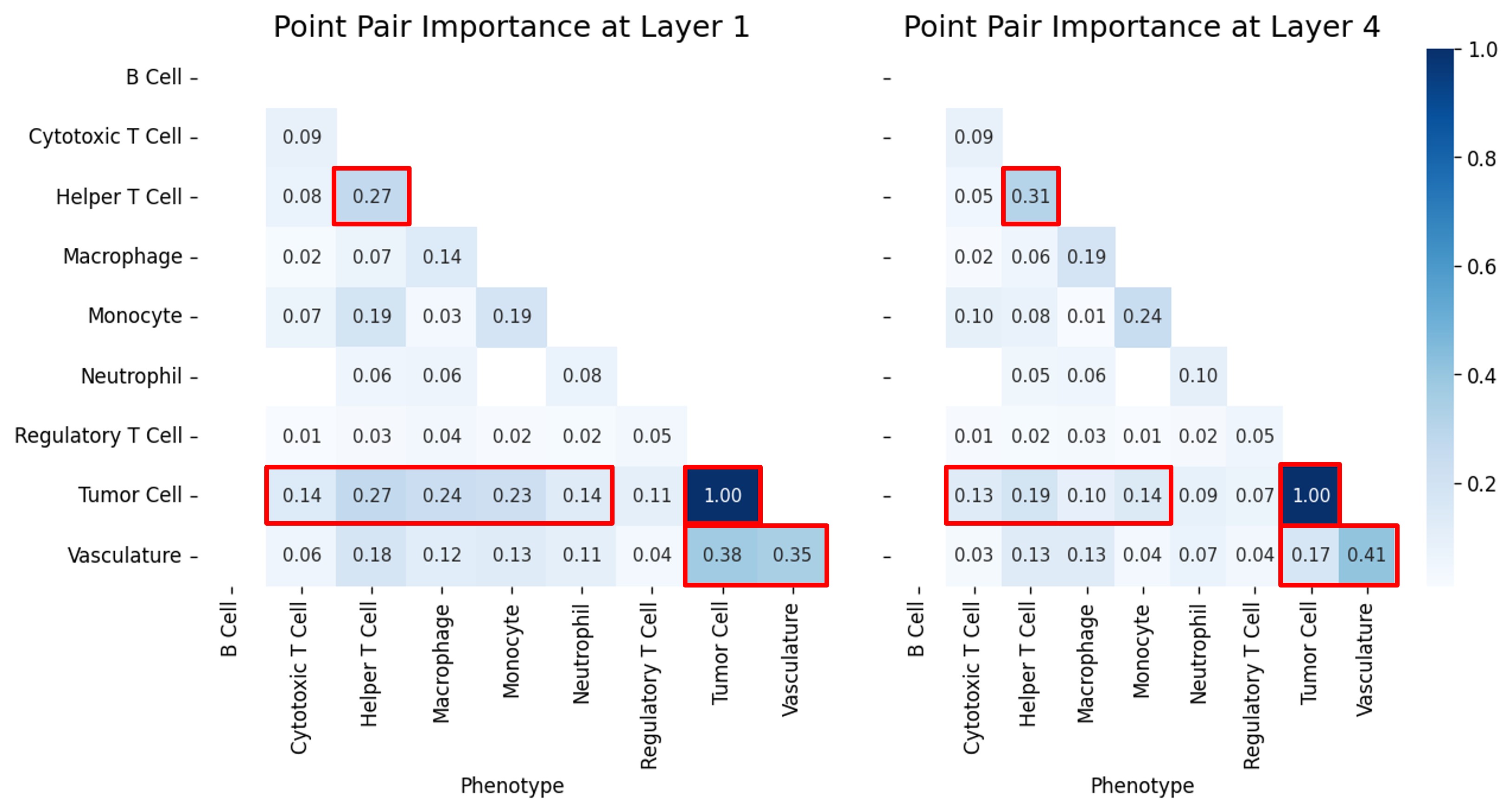}
    \caption{The relative importance between distinct cells that were learned through different layers of the point pair prioritization sub-network. (A few cases have been highlighted in red, which is in line with established biology practice.)}
    \label{importance}
    \vspace*{-3\abovedisplayskip}
\end{figure}
\section{Conclusion and Future Work.}\label{conclusion_FW}
In this paper, we propose SAMCNet, a neural network architecture with local reference frame characterization and a point pair prioritization sub-network.  SAMCNet provides a promising way to help understand the spatial configuration of multi-category point patterns and most relevant N-way spatial relationships. Experimental evaluation shows that the proposed model outperforms existing DNN techniques.

In the future, we plan to investigate a dynamic local reference frame characterization layer to learn the spatial distribution of an embedded feature space between a given point and its neighbor. We also plan to identify a multi-category public benchmark dataset for a larger and broader evaluation of the proposed method. We plan to extend this work to consider spatial variability by learning point pair importance based on density and distribution of multi-category points in different sub-regions.  
\section*{Acknowledgments.}
This material is based upon work supported by the NSF under Grants No. 2040459, 1737633, 1901099, 1218168, and 1916518. We also thank Kim Koffolt and the Spatial Computing Research Group for valuable comments and refinements. 

\bibliographystyle{ACM-Reference-Format}
\bibliography{bib}
\clearpage

\appendix

\section{Reproducibility}
\subsection{Code and Implementation Details}
To promote open science and reproducibility, the code 
used in the experiments are shared through Github \footnote{https://github.com/majid-farhadloo/SAMCNet_2022}. 

Patient privacy and the propriety nature of the data prevent us from publishing the dataset. But we would like to bring to the attention of the researcher community to evaluate the capability of the proposed model; a recent paper (Astropath) highlighted a similar dataset used in the experiments on big data for cancer immunology and conducting spatially-preserved analysis \cite{berry2021analysis}. However, this dataset was not publicly available at the time of submission. A more detailed discussion on this can be found here: https://ventures.jhu.edu/news/astronomy-pathology-biopath-biomarkers-cancer/. 

Table \ref{hyperparameter} presents the details of all parameters is used to train SAMCNet. 

\begin{figure}[H]
	\centering
	\begin{subfigure}{.3\textwidth}
	\centering
		\includegraphics[width=\textwidth]{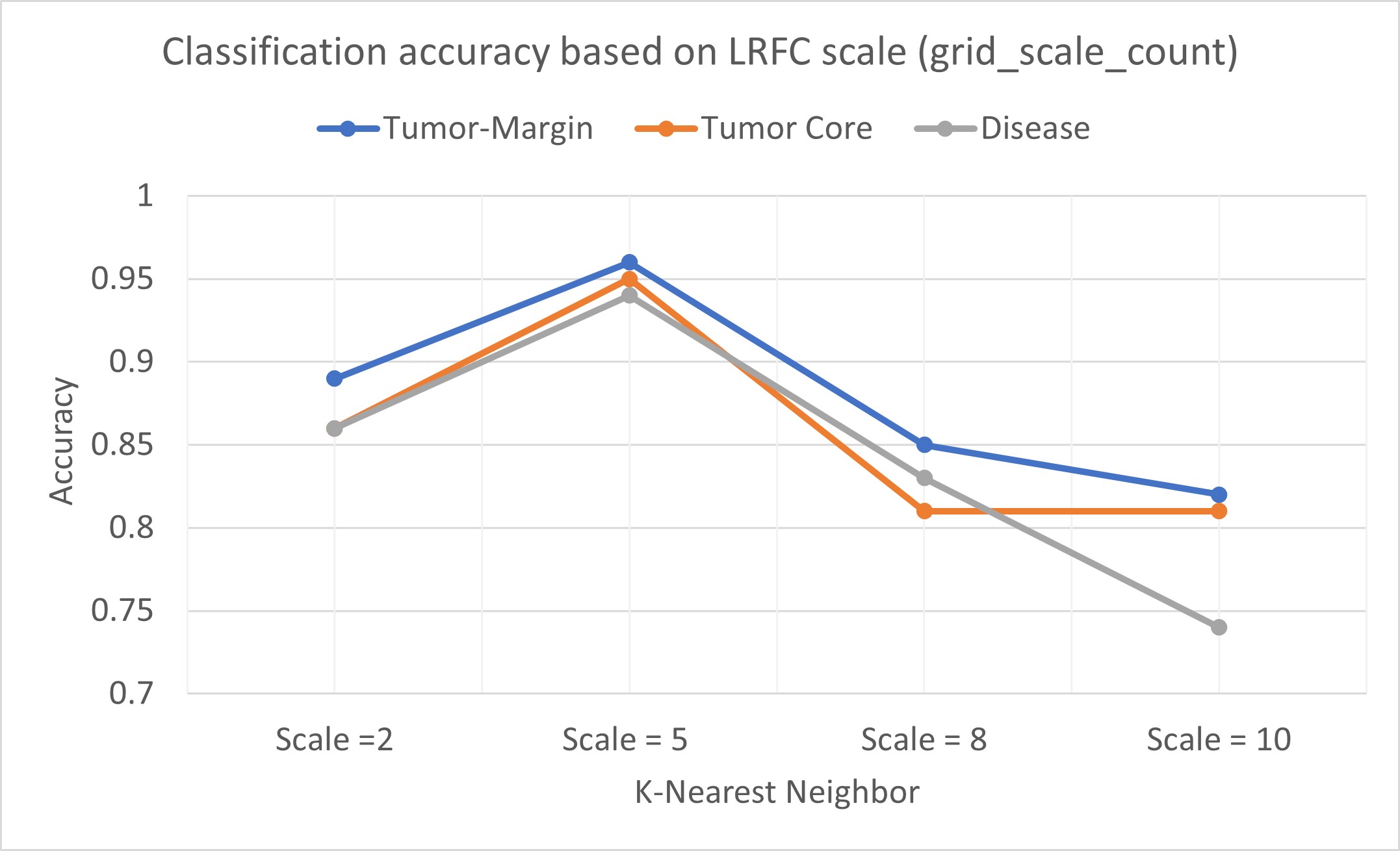}
		\caption{Grid-Scale count}
		\label{grid_scale}
	\end{subfigure}
	\begin{subfigure}{.3\textwidth}
	\centering
		\includegraphics[width=\textwidth]{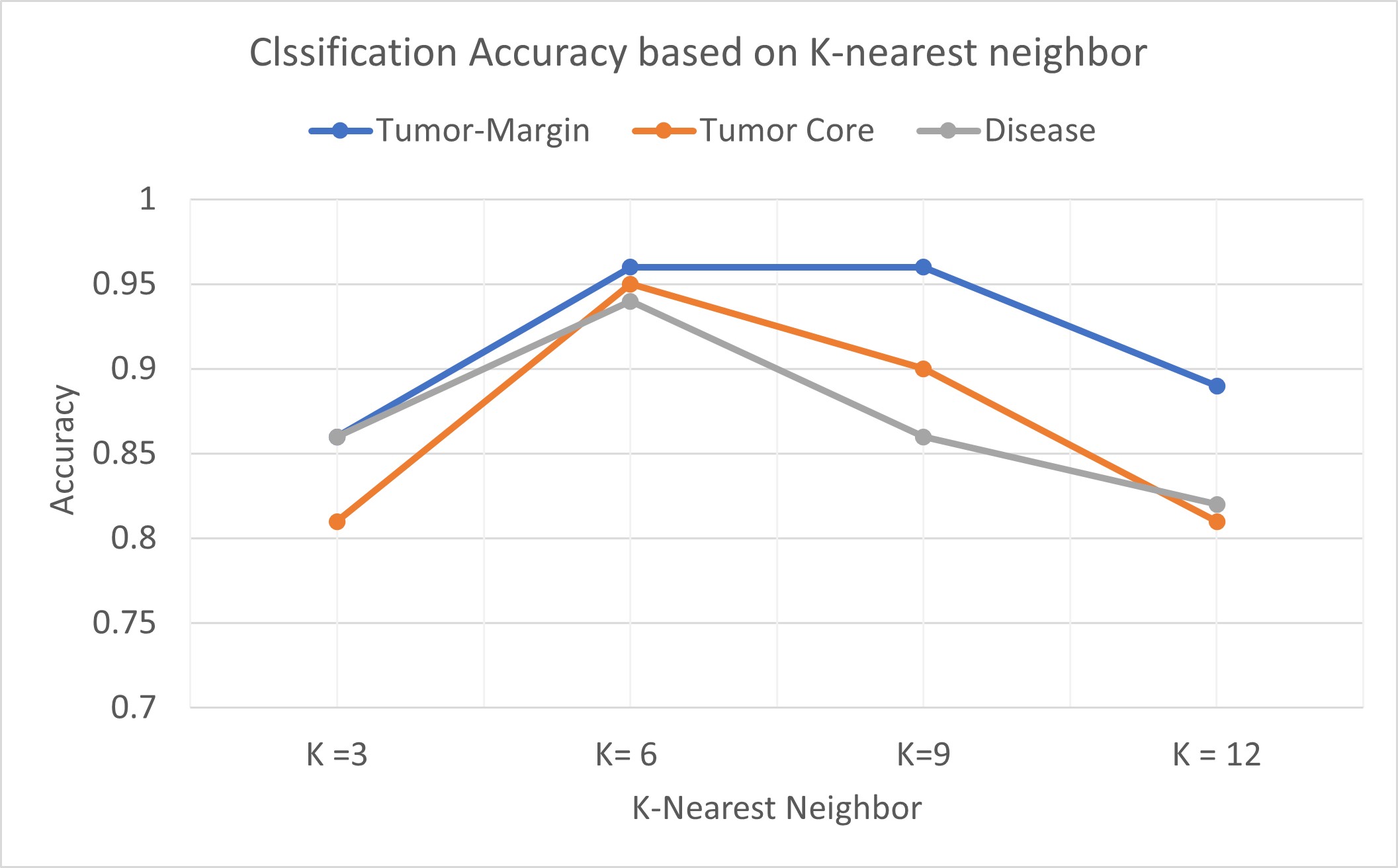}
		\caption{Number of nearest neighbors, k}
		\label{kneighbor}
	\end{subfigure}
	\begin{subfigure}{.3\textwidth}
	\centering
		\includegraphics[width=\textwidth]{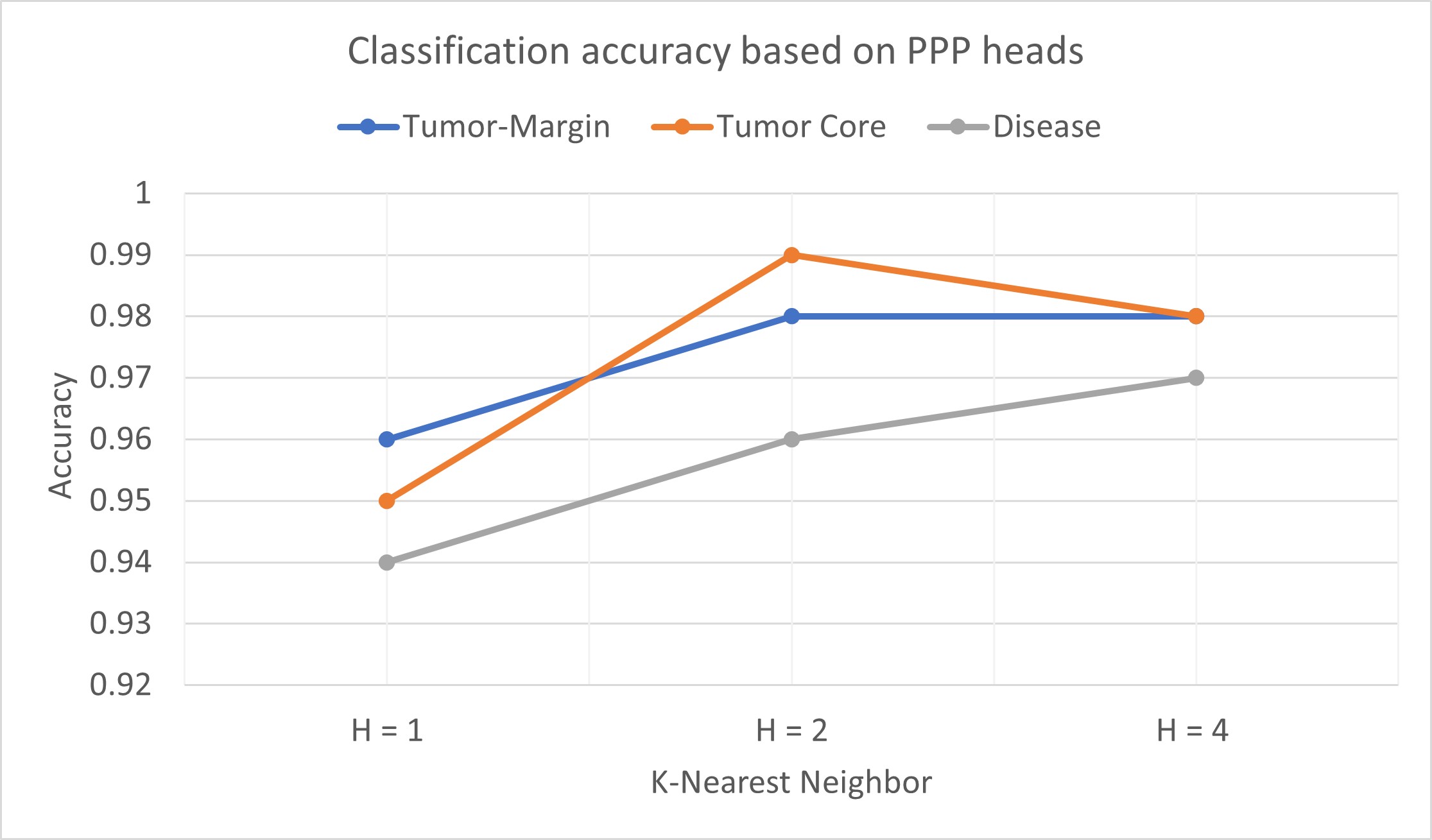}
		\caption{Number of heads in prioritization sub-network.}
		\label{heads}
	\end{subfigure}
	\caption{Key parameters evaluation.}
\end{figure}

\section{Additional Experiments}\label{more_sensitivity}
We further evaluated our proposed model by varying the key parameters, namely grid scale count, k-nearest neighborhood, and prioritization heads.

As shown in Fig. \ref{grid_scale}, the trends show that classification accuracy is sensitive to the choice of scale representation, where increasing grid-scale does not guarantee better performance. For example, the classification accuracy drastically dropped to lower than 0.75\% in disease classification when the grid-scale count (i.e., \textbf{\textit{s}} in equation \ref{scaling}) for a multi-scale representation was set to 10. While we did not thoroughly test all feasible grid-scale counts, our intuition is representing the relative distance in an embedded space very close or even higher than that first multi-layer perceptron (i.e., first EdgeConv in layer-1); makes it challenging to approximate local information. 
\begin{table}
\centering
\caption{Details of the parameter settings in proposed model.}
\begin{tabular}{|l|l|} 
\hline
Parameters & Value \\  \hline
Epoch       & 200    \\  \hline
Batch Size  & 7      \\ \hline
Learning rate  & 0.001  \\ \hline
momentum    & 0.9    \\  \hline
num\_points & 1024   \\  \hline
\begin{tabular}[c]{@{}l@{}}num\_heads\\(prioritization head)\end{tabular} & 1  \\ \hline
dropout     & 0.5    \\ \hline
k (nearest neighbor) & 6      \\ \hline
\begin{tabular}[c]{@{}l@{}}emb\_dims \\(shared fully-connected layer to \\aggregate multi-scale features)\end{tabular} & 1024   \\ \hline
min\_grid\_scale  & 1      \\  \hline
max\_grid\_scale  & 100    \\ \hline
grid\_scale\_count   & 5  \\ \hline
\end{tabular}
\label{hyperparameter}
\end{table}

We tested our model with different sizes of $k$ nearest neighborhoods. As shown in Fig. \ref{kneighbor}, a large k neighbor size results in deteriorating the classification performance. It shows that beyond a certain threshold density, the locally connected neighborhood graph fails to approximate geodesic distance and destroys the geometry of each patch, as discussed in \cite{wang2019dynamic}. In addition, this confirms the hypothesis that a large size of $k$ allows an overpowering amount of weak point pair interactions contributing to the overall representation of the center node. Hence, as discussed in Section \ref{attention_network}, one may investigate more in a combination of different aggregation operations as the size of K increases. 

Lastly, We also tested SAMCNet with a different number of  heads, $H = \{1, 2, 4\}$, at each layer. As shown in Fig. \ref{heads},  our point pair convincingly learned point pair interactions between various categorical attributes. This also implies that compared to a multi-head attention network, our one-head prioritization sub-network provides the best trade-off in model complexity, computational time complexity, and classification performance in terms of learning fewer parameters and taking less time.

\section{DNN Architectures for Point Pattern Classification}\label{other_related_works}

The success of convolutional neural networks (CNNs) in many pattern recognition tasks (e.g., \cite{farhadloo2019twitter, cecotti2020grape}) has inspired researchers to generalize convolution-like operations to directly apply to 2D/3D point cloud data without any computationally expensive intermediate conversion layers. PointNet \cite{qi2017pointnet}, the first neural network architecture that directly applies to point cloud data, learns point features independently through several fully connected neural network layers and aggregates them using an asymmetric function operation (e.g., Max pooling). PointNet++ \cite{qi2017pointnet++}, a variation of PointNet accounts for the local structure by applying graph coarsening operation and a shared PointNet recursively to a set of local points chosen by farthest point sampling and subsequently their k-nearest neighbors. However, these techniques are limited in learning fine-grained local structures since learning representation of each point independently at a localized scale to preserve permutation invariance.

DGCNN \cite{wang2019dynamic} proposes a graph dynamic graph CNN that dynamically updates the graph network at each layer by learning both local and global information.  This work is inspired by PointNet using a simple operation known as EdgeConv, where rather than independently applying to individual points, construct a locally connected neighborhood graph to exploit from both center nodes and edge features. Many other efforts have been made to learn local structure. For example, SpiderCNN \cite{xu2018spidercnn} proposes a multi-scale hierarchical that extend convolutional operations from regular grids to irregular point sets that can be embedded to ${\rm I\!R}^n$. 

However, these approaches are not designed to learn spatial relationships in multi-categorical point sets. In addition, they do not fully exploit the spatial distribution of points beyond simply measuring relative distance or applying a  discretization or feed-forward neural network to coordinates.


\section{Background: Spatial Co-location Measures.}
\textbf{Cross-K Function:} Spatial statistics \cite{intro_SDM} uses the cross-K function, a generalization of  Ripley’s K function, to detect spatial relationships between point patterns with more than one feature. The cross-K function $k(h)$ for binary spatial features is defined as $K_{ij}(h) =\lambda_{j}^{-1}E|\mbox{\# type $j$ instances within distance $h$}\\ \mbox{of a randomly chosen type $i$ instance}|$, where $i$ and $j$ represents two category types, $\lambda_j$ is the density of type $j$ instances, $h$ is the distance, and $E|$.$|$ is the expectation. The cross-k function could be estimated in the form of $K_{ij}^{\hat{}}(h) = \frac{1}{\lambda_i\lambda_j W} \Sigma_k\Sigma_l I_h(d(i_k, j_l))$, where $d(i_k, j_l)$ is the distance between the $i_k$ instance and the $j_l$ instance, $I_h$ is an indicator function, and $W$ is the study area \cite{shekhar2001discovering}.  The value of cross-k is a function of neighborhood distance $h$, which implies the spatial relationship between categorical points at different scales.

\textbf{Participation Index:} The co-location pattern interest measure most related to the cross-k function is the participation index. The participation index, an upper-bound approximation of the cross-K function, possesses an anti-monotone property that can be used for computational efficiency. Before defining participation index, we need to define another interest measure, participation ratio. The participation ratio $Pr(C, f_i)$ of feature $f_i$ in a co-location pattern C = $\{f_1, ..., f_K\},$ $1 \leq i \leq k$, is the fraction of spatial objects of feature $f_i$ in the neighborhood of instances of co-location C. Then, participation index $Pi(C)$ is defined as the minimum participation ratio of the features in a co-location pattern, that is $Pi(C) = min_{f_i \in C}\{Pr(C, f_i)\}$. 


The overall formulation of participation ratio is as follows: 

\begin{equation} \label{participation_ratio}
Pr(C, f_i) = \frac{\textrm{Number of distinct $f_i$ in instances of C}}{\textrm{Number of $f_i$}}\end{equation}

From equation above, it can be observed that the value of the participation index is between 0 and 1.  A  large $Pi(C)$  value shows that events of $f_i$ tend to be located in close spatial proximity of other events of features in C. We used the cross-k function and participation index to quantify the spatial relationship between categorical point sets. For example, given a point set containing points belonging to $g$ categories and a set of neighborhood distance threshold $H = \{h_1, ..., h_s\}$, $1 \leq i \leq s$, there will be $g(g-1) * s$ cross-k functions or participation index pairs.

\section{LRFC}
Local reference frame characterization is proposed based on the following theorem which proof is given in \cite{gao2018learning}. 
\begin{theorem}
    Let $\Psi(x) = (e^{i\langle a_j,x\rangle}, j=1,2,3)^T \in \mathbb{C}^3$ where $e^{i\theta}=\cos\theta + i\sin\theta$ and $\langle a_j,x\rangle$ is the inner product of $a_j$ and $x$. $a_1,a_2,a_3 \in \mathbb{R}^2$ are 2D vectors such that the angle between each pair is $2\pi/3, \forall j, \| a_j \| = 2\sqrt{\alpha}$. Let $C \in \mathbb{C}^{3\times 3}$ be a random complex matrix such as $C*C=I$. Then $\phi(x)=C\Psi(x)$, $M(\Delta x)=C diag(\Psi(\Delta x))C*$ satisfies
    \begin{equation}
        \phi(x+ \Delta x) = M(\Delta x)\phi(x)
    \end{equation}
    and
    \begin{equation}
        \langle \phi(x+ \Delta x), \phi(x) \rangle = d(1-\alpha \|\Delta x\|^2)
    \end{equation}
    where $\phi(x)$ is the representation of location $x$, $d=3$ is the dimension of $\phi(x)$, and $\Delta x$ is a small displacement from $x$. \label{theorem:loc_representation}
\end{theorem}$\Psi(x)$ is represented as a concatenation of the position embedding ($PE$) at $S$ scales. 

\end{document}